# Sample Size Calculations for Developing Clinical Prediction Models: Overview and pmsims R package

D. Shamsutdinova[1,2,†], F. Zimmer[1], O. R. Olaniran[1,2], S. Markham[1], D. Stahl[1,2], G. Forbes*[1,2], E. Carr*[1]

[1]*Department of Biostatistics and Health Informatics, King's College London*
[2]*NIHR Biomedical Research Centre, Maudsley NHS Trust*

## Abstract

**Background:** Clinical prediction models are increasingly used to inform healthcare decisions, but determining the minimum sample size for their development remains a critical and unresolved challenge. Inadequate sample sizes can lead to overfitting, poor generalisability, and biased predictions. Existing approaches—such as heuristic rules, closed-form formulas, and simulation-based methods—vary in flexibility and accuracy, particularly for complex data structures and machine learning models.

**Methods**: We review current methodologies for sample size estimation in prediction modelling and introduce a conceptual framework that distinguishes between mean-based and assurance-based criteria. Building on this, we propose a novel simulation-based approach that integrates learning curves, Gaussian Process optimisation, and assurance principles to identify sample sizes that achieve target performance with high probability. This approach is implemented in *pmsims*, an open-source, model-agnostic R package.

**Results**: Through case studies, we demonstrate that sample size estimates vary substantially across methods, performance metrics, and modelling strategies. Compared to existing tools, *pmsims* provides flexible, efficient, and interpretable solutions that accommodate diverse models and user-defined metrics while explicitly accounting for variability in model performance.

**Conclusions**: Our framework and software advance sample size methodology for clinical prediction modelling by combining flexibility with computational efficiency. Future work should extend these methods to hierarchical and multimodal data, incorporate fairness and stability metrics, and address challenges such as missing data and complex dependency structures.

*Abbreviations:* AI – artificial intelligence, ML - machine learning, MAPE – mean absolute prediction error, RF- random forest, SVM – support vector machines, ICI – integrated calibration index, GP – gaussian processes

*Joint last authors
† Corresponding author; diana.shamsutdinova@kcl.ac.uk

# 1. Introduction

Clinical prediction models are increasingly used to support healthcare decision-making by estimating the probability of various health outcomes from patient characteristics. These models, whether statistical or machine learning-based, require sufficient data to ensure reliable performance. However, determining the appropriate sample size for developing such models remains a complex and often-overlooked challenge (Damen et al., 2016; Dhiman et al., 2023). In traditional hypothesis testing, sample size is based on power and precision for a specific parameter. In prediction modelling, sample size considerations focus on generalisability and control of overfitting and optimism, as well as on ensuring the stability and reliability of individual predictions. Sample size requirements for prediction models can be driven by model and data characteristics, including the outcome distribution, number of predictors, noise-to-predictors ratio, strength of predictor-outcome relationships, model complexity, presence of non-linear and interaction terms, and the acceptable level of performance (Austin & Steyerberg, 2017; Kalaycıoğlu et al., 2025; Pavlou et al., 2024; Riley et al., 2019a, 2019b).

Existing methods to estimate minimum sample sizes for prediction model development include simple heuristics, closed-form analytic solutions, and simulation-based approaches. Heuristics, such as the 10-20 events per predictor variable rule (Austin & Steyerberg, 2017; Peduzzi et al., 1996), provide quick but oversimplified guidance that only captures generic data features. Closed-form methods (Pavlou, 2021; Riley et al., 2019a) offer fast, interpretable formulas under distributional assumptions but are not readily extendable to complex data structures or alternative modelling frameworks. Simulation-based approaches provide the greatest flexibility for complex data and machine learning models at the cost of computational burden (Goldenholz et al., 2023; Pavlou et al., 2025; Riley, Whittle, et al., 2025).

Although some of the sample size calculation methods have been implemented in software packages, including *pmsampsize* (Ensor et al., 2022) and *samplesizedev* (Pavlou et al., 2025), or published as formulas (Silvey & Liu, 2024), many lack practical implementations for researchers, limiting their uptake and utility. However, despite advances in methodology and the introduction of reporting guidelines (e.g., TRIPOD-AI (Collins et al., 2024) most published models are still developed with inadequate sample sizes, risking overfitting and biased predictions (Damen et al., 2016; Dhiman et al., 2023). This underscores both the scarcity of existing software for sample size calculations (Ensor et al., 2022; Pavlou, 2021) and the limited engagement from researchers, highlighting the need for further development of user-friendly tools adaptable to diverse prediction algorithms and data modalities.

This paper addresses these gaps by reviewing existing approaches, setting out a theoretical framework for sample size estimation in clinical prediction modelling, and proposing a new simulation-based approach *pmsims* for calculating sample sizes (Carr et al., 2025). Our approach differs from previous solutions by combining learning curves, surrogate modelling, and Gaussian process optimisation to improve efficiency, and by identifying sample sizes that ensure adequate performance with high probability ('assurance'), rather than only on average. The pipeline is implemented in *pmsims*, a flexible, user-friendly, model-agnostic software tool for applied researchers.

## 2. The problem of sample size in clinical prediction modelling

### 2.1 The problem setting

We consider the task of developing a prediction model using a set of candidate predictors and a development sample drawn from a given population. Although we focus on binary outcomes, the problem setting extends naturally to continuous and time-to-event outcomes. The sample size problem is to determine the sufficient development sample for the model to generalise beyond the development sample and achieve adequate performance in the target population.

**Data-generating distribution.** The study population is defined by the joint distribution of predictors $x = (x_1, \dots, x_p)$, and the outcome $y$, denoted as $P_{gen} = P(x_1, \dots, x_p, y)$. In binary case, $y \in {0,1}$ is a Bernoulli random variable, characterised by the conditional probability $\pi(x) = P(y = 1 | x)$. Note that predictors determine probability $\pi(x)$, while individual outcomes $y$ are drawn from this probability, leading to irreducible prediction error even for perfectly specified models.

**Prediction model.** A prediction model is a function $f(x, w)$ from a model class $H$ parametrised by w. Model parameters are estimated by minimising a loss function on the development data. For logistic regression,

*Equation 2*

$$f(x, w) = \frac{1}{1+e^{-w^\top x}},$$

with parameters estimated by minimising the negative log-likelihood.

**Training and validation setup.** Given a development sample of size $n$, $D_n = \{(x_{train}^i, y_{train}^i)_{i=1..n}\}$, $D_n \sim P_{gen}{}^n$. The fitted model is $\hat{f}(x, w) = f(x, \hat{w}(x_{train}, y_{train}))$. Performance is typically evaluated on an independent validation dataset to approximate how well the model generalises to the population.

## 2.2 Sample size and model performance

Assume that we fitted a prediction model $\hat{f}$ to a development set $D_n$ and chose a performance measure $M$ (e.g., AUC, calibration slope, Brier score for binary, or mean absolute prediction error (MAPE), $R^2$ for continuous outcomes (Steyerberg, 2019)). The model's expected performance in the underlying population will be $E(M|D_n) = E_{P_{gen}}(M \mid D_n)$.

Before the developmental data is collected, the expected performance would be averaging the above over possible drawings of the development sets of size $n$:

*Equation 3*

$$E[M|n] = E_{D_n}[E(M|D_n)].$$

**Correct versus mis-specified models.** A central assumption in the sample size problem is that the expected performance improves as the sample size increases. Indeed, for correctly specified models, the expected predictive performance converges to the optimal within the model class $H$ level, as $n \to \infty$ in many settings.

Correct specification means the model class $H$ contains the true $\pi(x)$. The quality of the parameters' estimates typically improves as $n \to \infty$, that is, $\hat{w}(n) \xrightarrow[n \to \infty]{} w_{true}$, so

*Equation 4*

$$\hat{f}(x) \xrightarrow[n \to \infty]{} f^*(x) = \pi(x),$$

and the performance reaches its optimal level. For example, this is the consistency result for maximum likelihood estimators in correctly specified models (Wald, 1949), and empirically evidenced for complex models such as neural networks (Kaplan et al., 2020).

In practice, the true model is unknown. If the model is mis-specified, the asymptotic predictive performance may differ from theoretical expectations and never reach the optimal level (Kalaycıoğlu et al., 2025).

## 2.3 Two approaches to the sample size problem: mean and assurance.

Two formulations of the sample size problem can be distinguished:

**Mean-based criterion.** Find the smallest $n$ such that the expected performance exceeds a target level $M^*$:

$$min\ n: E_{D_n} E(M|D_n) > M^*$$

*Equation 5*

**Assurance criterion.** This more stringent formulation requires that the performance exceeds $M^*$ with a high probability, e.g. 80%. This ensures that most models trained on datasets of size $n$ will reach the desired performance, explicitly accounting for variability across development sets.

$$min\ n: P_{D_n}(E(M|D_n) > M^*) > \delta.$$

*Equation 6*

The mean and assurance criteria are derived by treating the model's performance $E(M|D_n)$, based on a specific dataset $D$ of size $n$, as a random variable with respect to the dataset sampled. This variable is described by the distribution of the model performance in the target population across all draws of development data of size $n$. The goal is then to find the smallest $n$ such that either the mean or a chosen quantile exceeds the target performance $M^*$; $\delta=0.8$ corresponds to the 20$^{th}$ percentile exceeding $M^*$. For instance, under the mean-based criterion, fitting a prediction model to 100 random datasets of size $n$, would result in some models performing above, some below $M^*$, with the mean above $M^*$. In the second approach, a fixed proportion of the models outperforms $M^*$. This also illustrates that the assurance criterion goes beyond the mean of the distribution of $E(M|D_n)$, and considers model variance. This implies, for example, that models where estimated parameters vary considerably with small changes in the development data (e.g., deep neural networks) would require larger development data to achieve good performance with certainty (**Figure 1**).

**Figure 1. Two approaches to the sample size problem: mean and assurance.**

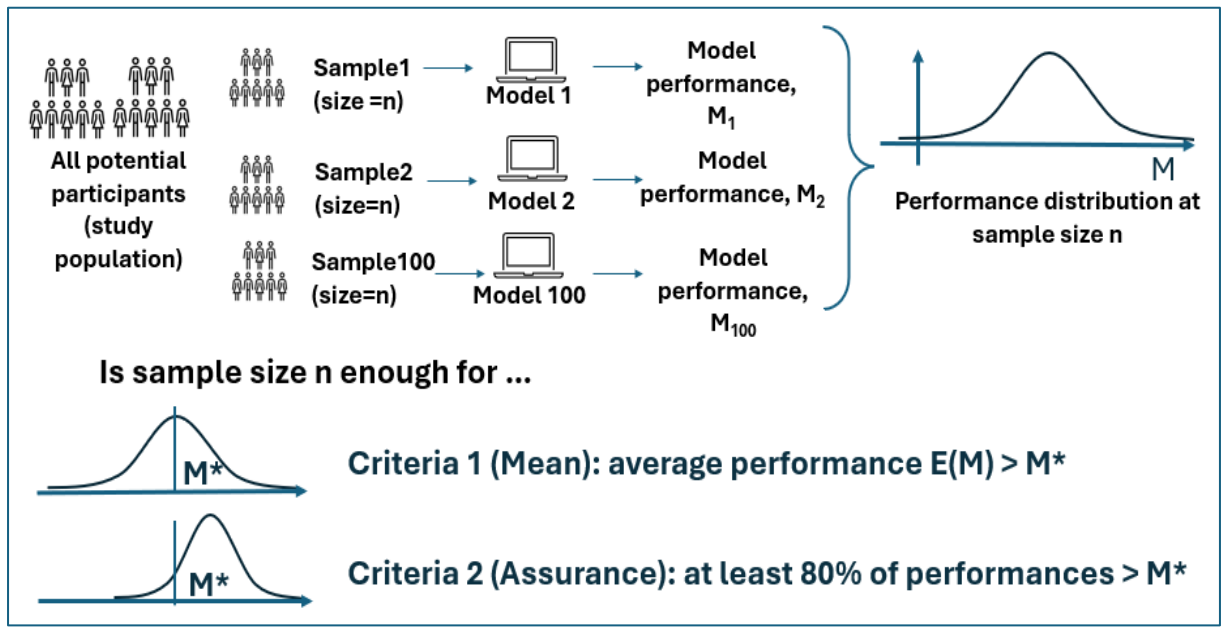

## 2.4 Sources of randomness

The performance of a predictive model trained on a finite sample is a random quantity, influenced by several sources of variability. Understanding which of these can be mitigated by increasing the sample size is central to the sample size problem.

1) **Irreducible outcome error.** Even if the true conditional distribution $P(y \mid x)$ were known, the realised outcomes contain intrinsic randomness that cannot be eliminated, reflecting biological or measurement variability. That is, for binary outcomes, even a perfect model cannot predict $y$ without error when $0 < \pi(x) < 1$. This error represents the lower bound of prediction error and is not reducible by increasing the sample size.

2) **Development data variability.** Random samples drawn from the same population yield different prediction models. Smaller samples tend to generalise poorly and result in overfitted models. Sample characteristics (means, variances, correlations between the features) vary more across small samples than across large ones, hence models developed using small samples show higher instability. This source of variability decreases with larger samples as the fitted model converges to its population optimum. This is the key component addressed by sample size calculations.

3) **Model-fitting randomness.** Many machine learning algorithms (e.g., random forests, neural networks) introduce stochastic elements—random feature subsampling, weight initialisation, or optimisation paths. This leads to variability in the fitted model even for identical development data (Riley, Ensor, et al., 2025). It can be partially mitigated by averaging (ensembles, repeated development), but it is not fully eliminated by larger sample sizes.

4) **Validation data variability.** Estimated performance metrics depend on the finite validation sample. This variability is *not* reduced with larger *development data,* which may yield a more accurate model, but would not decrease the uncertainty of performance estimates obtained in an *independent* validation cohort. In internal validation procedures such as cross-validation, each validation fold consists of a fixed proportion of the development sample. Thus, enlarging the development dataset automatically increases the size of the validation subsets, which in turn improves the precision of internally validated performance estimates. In contrast, the required validation sample size depends on the desired precision of the performance metric and is a related but distinct design problem, outside the scope of this paper addressed elsewhere (Archer et al., 2021; Pavlou, 2021).

In simulation-based approaches, developmental sample and model-fitting randomness can be explicitly represented by repeatedly sampling development datasets and re-fitting models. Validation data variability is often mitigated by generating a large validation sample, independent from the development samples for accurate estimate of the model's true performance in the population. Irreducible outcome error is encoded in the data-generating mechanism. Thus, simulations capture both population-level uncertainty and model instability. However, they rely on correct specification of the data generator, and any biases arising from misspecification should be carefully considered.

## 2.5 Summary

The problem of sample size calculation for prediction modelling involves solving equations 5 or 6. As the required sample sizes can vary across performance metrics, a combined approach targeting multiple metrics, such as AUC and calibration slope, may be needed.

# 3. Overview of existing approaches

## 3.1 Recommendations based on events per variable

An early approach to sample size calculation was a heuristic rule of 10 events per predictor variable (EPV) (Peduzzi et al., 1996), referring to events rather than observations to account for outcome prevalence. For example, a model with 10 predictors would require >100 events. Although simplistic, it remains a common reference for feasibility. Steyerberg et al. showed that at least 10-20 EPV is needed to avoid overfitting (measured by shrinkage > 0.9) (Steyerberg, 2019), and 20 EPV for an accurate c-statistic estimate (Austin & Steyerberg, 2017). Others found 5 EPV adequate when predictors are weakly correlated and penalisation is applied, suggesting sample size depends more on signal-to-noise ratio than event count alone (Ogundimu et al., 2016; Vittinghoff & McCulloch, 2007). Overall, EPV rules oversimplify by ignoring predictor strength, correlations, and penalisation, motivating more principled approaches.

## 3.2 Closed-form analytic approaches

Theory-grounded analytical approaches derive sample size requirements using explicit formulas under simplifying modelling assumptions. An early example of closed-form analytical approaches for sample size calculations for prediction modelling can be found in Jinks et al., 2015 [25], which focuses on survival outcomes.

A central example approximating the development sample size for linear, logistic, or Cox regression models was provided by Riley et al. (Riley et al., 2019b, 2019a) and the corresponding R package *pmsampsize* (Ensor et al., 2022). Their solutions utilise asymptotic theory to derive performance measures that can be directly linked to sample size, building on earlier works (Van Houwelingen & Le Cessie, 1990). Accounting for the outcome prevalence, number of predictors, and expected model fit (e.g., $R^2$ for linear models, AUC for binary classification), the sample size should satisfy three conditions: (1) uniform shrinkage > 0.9 to limit overfitting, (2) reliable estimate of the overall fit, measured by the difference between the in- and out-of-sample $R^2$, and (3) accurate estimation of the mean risk (Pate et al., 2023). Typically, the shrinkage condition is most restrictive. Similar to Austin and Steyerberg (2021) (Austin et al., 2021), Riley et al.'s works (Riley et al., 2019a, 2019b) prioritise controlling overfitting rather than targeting absolute performance. The latter was considered by Van Smeden et al. (van Smeden et al., 2019) for discrimination, calibration, and stability of individual predictions.

Although closed-form solutions, and *pmsampsize* in particular, have been widely used and are efficient and user-friendly, they have limitations. Their reliance on distributional assumptions restricts generalisability, especially for complex settings such as multilevel data, high-dimensional predictors, or machine learning (ML) models. Empirical evaluations have shown that analytic methods may underestimate required sample sizes in certain scenarios (Kalaycıoğlu et al., 2025; Pavlou et al., 2024).

### 3.3 Simulation-based approaches

Simulation-based approaches directly address the limitations of closed-form methods by providing greater flexibility for representing realistic data-generating mechanisms and extending the choice of prediction models. Such methods explicitly define the data-generating process and estimate performance empirically through repeated sampling. This design enables exploration of settings where theoretical results are unavailable, such as non-linear feature effects, strong interactions, varying noise-to-signal ratios, and even intentional model misspecification (Austin et al., 2021; Kalaycıoğlu et al., 2025). The advantages of flexibility and realism come at a cost of computational intensity, coding complexity, and dependence on simulation assumptions. As a compromise, hybrid methods have been proposed, where estimated learning curves are summarised into approximate analytical formulas for practical application (Silvey & Liu, 2024; van Smeden et al., 2019)

Recent work by Riley et al. has formalised this flexibility into a general sample size framework (Riley, Whittle, et al., 2025). The approach defines key simulation inputs, including data distribution parameters (prevalences, dimensionality, expected c-statistic), and modelling strategy (classical or penalised regression, ML). The framework's key outcome is the posterior distribution of individual predictions and performance metrics, including calibration, discrimination, clinical utility, fairness, and prediction error, for a given sample size. Distributional output allows examining sample size requirements under the mean and assurance criteria. It further facilitates exploration model stability and uncertainty around individual predictions and population-level performances with respect to the development sample size.

The package *samplesizedev* (Pavlou et al., 2025) implements many elements of this framework and uses simulations to apply the assurance criteria for calibration slope, and reporting measures of variability for the simulated calibration slope, mean prediction error and AUC.

## 3.4 Data generation for simulation-based approaches

In the simulation-based approach, data generation is crucial because results depend on generating datasets of arbitrary size that reflect key characteristics of the target data. Data generators aim to reproduce essential data features, such as number and type of predictors, correlations, or nonlinear dependencies. Pilot samples are valuable sources for calibrating generators, either directly or by getting estimates for the key distributional parameters. Most existing approaches use simple parametric generators, such as independent Bernoulli or normally distributed predictors combined with generalised linear models for outcome simulation (*samplesizedev*). This is computationally efficient and useful when pilot data are unavailable, but it relies on strong distributional assumptions and often fails to capture complexities like heterogeneity, skewness, and multicollinearity.

Other studies anchored simulations in real data by fitting models to large medical datasets (n≈2000-200,000), replicating predictor-outcome relationships (Austin et al., 2024; Kalaycıoğlu et al., 2025; van der Ploeg et al., 2014) and regenerating outcomes from these models to explore data generators and prediction models influence sample size. Prediction models may be aligned with, or deliberately differ from, the data-generating mechanism to assess the impact of model misspecification. While rooting simulations into real data improves realism, predictors are not synthesised, limiting the investigation of predictor space and generalisability. Silvey & Liu (2024) (Silvey & Liu, 2024) partially addressed this by using 16 distinct health datasets (70,000-1,000,000) to simulate predictive performance across algorithms and sample sizes, mapping sample size requirements to predictor characteristics and data nonlinearity.

Summing up, current methods often fail to reflect real data complexity. Future integration of sophisticated generators that handle correlated and hierarchical data (Goldfeld & Wujciak-Jens, 2020; Nowok et al., 2016) and employ deep learning techniques (C. Lu et al., 2024) could further enhance accuracy.

## 3.5 Learning curve frameworks

A parallel line of work, originating from an ML perspective, conceptualises model performance as a function of sample size. This learning curve framework represents performance as

$$M = g(H, n)$$

where $H$ is the model class, and $g(.)$ is a continuous, typically monotonic curve mapping sample size to performance metrics such as AUC, mean prediction error, or accuracy (Figueroa et al., 2012; Provost et al., 1999; Silvey & Liu, 2024). Sample size estimation then amounts to approximating $g(.)$ and identifying the minimum $n$ that meets a pre-specified performance level. Estimation can be performed through direct simulation across sample sizes, or parametrically, for example, by fitting inverse power law models for interpolation and extrapolation (Viering & Loog, 2023). Extrapolation involves fitting learning curves to the pilot data, estimating shape parameters, and inferring $n$ for the target performance. Extrapolation shows variable accuracy (Figueroa et al., 2012), and better results were achieved by enriching available samples with similar external data (Dayimu et al., 2024).

### 3.6 Gaussian process methods for simulation studies

Gaussian process (GP) surrogate methods have been proposed to reduce the computational burden in simulation studies (Zimmer et al., 2023). GPs enable smooth interpolation of the learning curve, provide robust confidence intervals, and can substantially reduce computation by focusing simulations on sample size regions most likely to contain the solution (Zimmer & Debelak, 2025). Because GPs estimate both the mean and deviation of target parameters, they can be integrated into Bayesian frameworks, as demonstrated in Dayimu et al. (2024) (Dayimu et al., 2024), where they used the posterior learning curve estimates as priors for pilot data to improve sample size estimates.

### 3.7 Summary

Overall, the transition from EPV heuristics to analytic and simulation-based approaches reflects a shift toward more flexible, model-agnostic, and empirically grounded methodologies. These developments have laid the foundation for modern techniques, including the *pmsims* package proposed here, which utilises simulation-based and learning-curve principles to estimate the minimum sample size required to achieve a desired level of predictive performance with assurance.

# 4. pmsims: a simulation-based, model-agnostic approach

The *pmsims* package estimates the minimum sample size required to achieve a desired level of predictive performance with high probability. It is fully model-agnostic, relying on user-defined components that specify how data are generated, how models are fitted, and how performance is measured. The workflow comprises four main steps:

**Step 1. Defining the simulation scenario by specifying the following inputs.**

a. **Data generator**: user-defined or standard. Standard generators allow choosing the outcome type (binary, continuous, or time-to-event) and number, type, and distributions of predictors (including proportions of true vs noise predictors).
b. **Model function:** the prediction model to be fitted (standard options or a user-defined function; supports regression, penalised models, and ML).
c. **Metric function:** the performance metric(s) (e.g., AUC, calibration slope, or user-defined).
d. **Expected and acceptable model performances:** user must specify the expected 'large sample' performance $M^{ideal}$, and the acceptable deviation $d$ from $M^{ideal}$. This defines the acceptable performance $M^* = M^{ideal} - d$. The sample size is solved for the model to achieve $M^*$ with the chosen assurance. A default assurance level is 80%.

**Step 2. Tuning the data generator.** To reflect the characteristics of the target population and provide a realistic basis for simulation, tune the data generator to (a) match the simulation parameters specified in step 1, and (b) ensure that the prediction model achieves the specified performance $M^{ideal}$ on large samples.

**Step 3. Estimating the learning curve.** Generate synthetic datasets of varying $n$, repeatedly fit the chosen model, and assess performance on independently simulated test data. The algorithm minimises the difference between the $M^*$ and the 20th quantile of the learning curve at sample size $n$, $M(n)$. The process is optimised using techniques such as bisection and Gaussian-process regression, as implemented in *mlpwr* (Zimmer et al., 2023) by allocating computational budget to learning curve region where solution is likely to be located.

**Step 4. Determining the minimum sample size.** Define the minimum $n$ as the smallest sample size for which the 20th percentile of the test performance distribution exceeds the pre-specified threshold. This ensures the model will meet the desired performance $M^*$ in at least 80% of simulated cases, providing an interpretable assurance criterion for model generalisability.

This framework combines the flexibility of simulation-based methods with the efficiency of surrogate learning-curve modelling, making *pmsims* suitable for a wide range of prediction modelling settings. Further practical information on package deployment, as well as the source code can be found at https://pmsims-package.github.io/pmsims/ (Carr et al., 2025).

## Figure 2. The pmsims workflow.

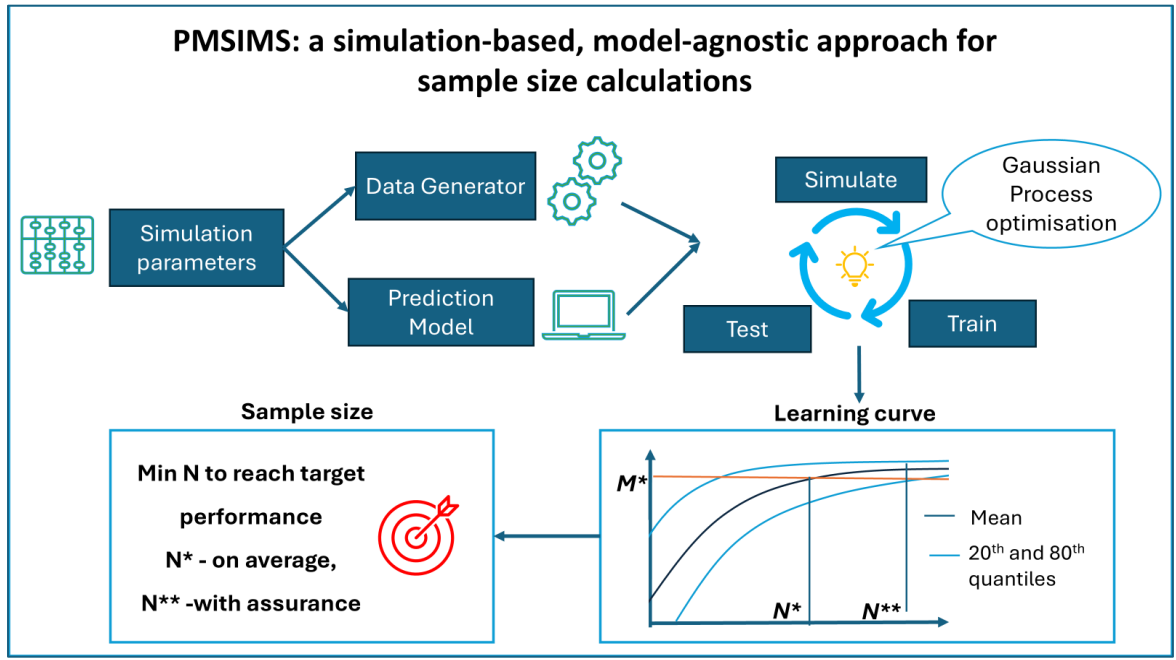

The figure presents the internal structure of the *pmsims* package. First, the simulation parameters are given by a user, including number of the true and noise predictors, outcome type (binary/continuous/time-to-event), type of the prediction model, and a performance metric. These parameters configure the data generator and prediction model blocks. The simulations are then performed to iteratively generate the development data of various sizes, fit prediction model and validate its performance on an independently generated test sample. The sample size is then inferred from the estimated learning curve.

## 4.2 Case study: illustrative implementation of *pmsims* and other methods

To illustrate the implementation of *pmsims* and other methods, we applied them to three case studies. The aim of this case study was to illustrate the variability in sample size estimates a researcher developing a prediction model may get using various approaches, rather than to justify a specific number.

Sample sizes were computed using 1) *pmsampsize*; 2) *samplesizedev* (Ensor et al., 2022; Pavlou et al., 2025); 3) *pmsims*; 4) the Shiny app from Silvey & Liu (2024) (Silvey & Liu, 2024); 5) the formula from Riley et al., 2020 (Riley et al., 2020); 6) results of Kalaycıoğlu et al. (2025) (Kalaycıoğlu et al., 2025); 7) 10 EPV heuristic; 8) combined results of Ploeg et al. (2014) and Austin et al. (2024) (Austin et al., 2024; van der Ploeg et al., 2014). All methods were implemented using their recommended criteria. The three case studies (0.063 prevalence, 10 predictors, 0.82 AUC; 0.11 prevalence, 44 predictors, 0.86 AUC; 0.25 prevalence, 17 predictors, 0.80 AUC) represented the datasets used by Kalaycıoğlu et al. (case 1 and 2) and by Silvey & Liu (case 3). All methodologies accommodated logistic regression, several supported ML. For illustration, sample sizes for ML models were pooled into a single category. All assumptions, R code and detailed results can be found **Supplementary Materials,** https://github.com/dianashams/pmsims_paper/tree/main.

Results show that minimum sample sizes vary considerably depending on the prediction targets (AUC, MAPE, calibration), model type, methodology, and the alignment between the true and assumed predictor-outcome relationship (**Figure 3**). For logistic regression, sample sizes required by various methodologies were between 200 and 6000 for Case 1, 1000 and 15,000 for Case 2 and Case 3, and significantly higher if misspecified (20,000+). ML models required considerably larger development datasets (5-10x on average) and exhibited greater diversity between models and methodologies, with estimates ranging from 2000 to tens of thousands.

The *pmsims* estimates (black triangles in **Figure 3**) for achieving a calibration slope of 0.90 with assurance were 3510, 4198, and 1439 for the three cases, in the middle of the overall ranges for the cases, and comparable to *samplesizedev* estimates of 940, 5047, and 1857 for the same calibration slope criterion (grey triangles**)**.

# Figure 3. Estimated sample sizes by different methodologies for three case studies

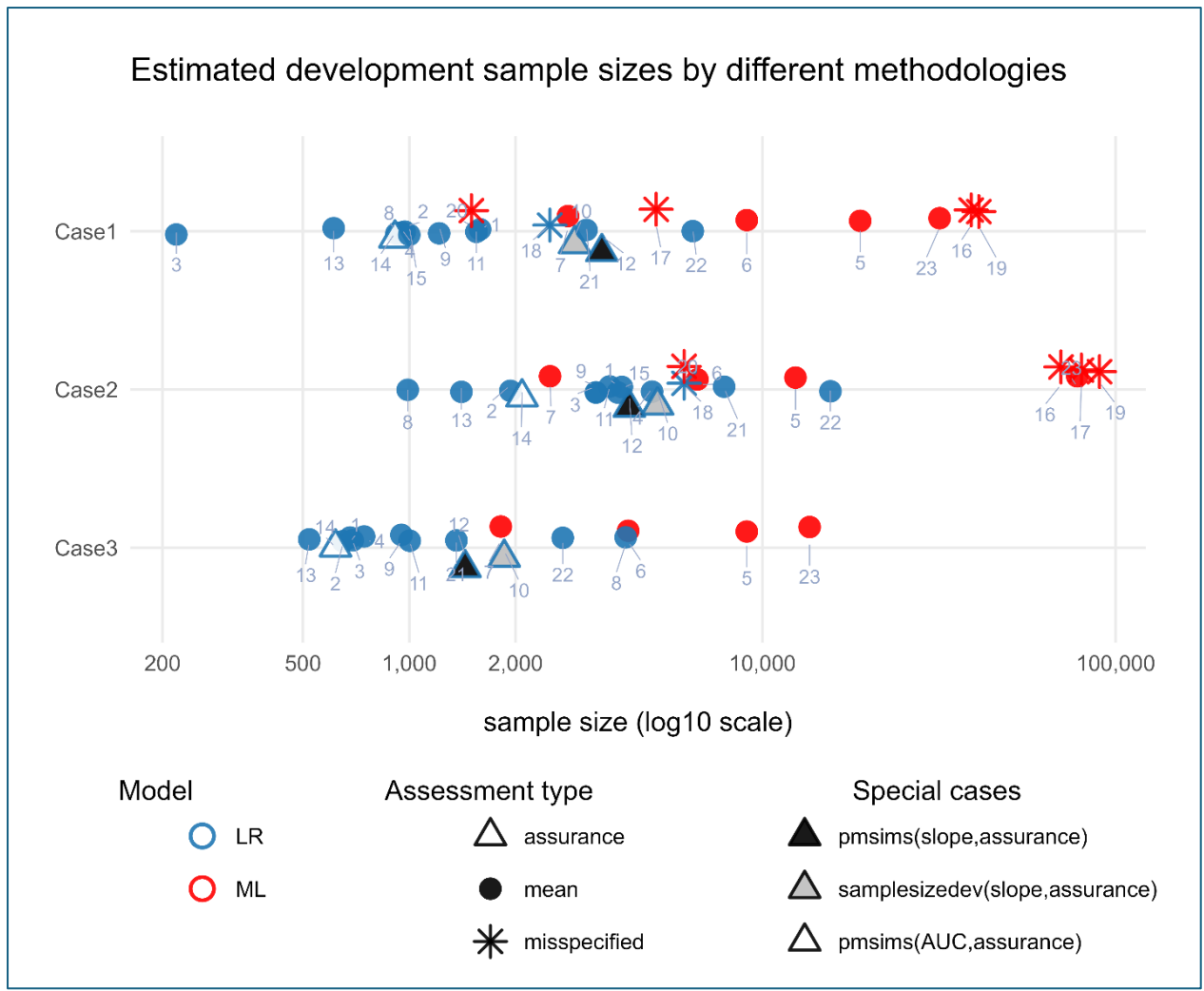

The sample sizes were computed using 1) heuristic 10 EPV (1 estimate, #1); 2) *pmsampsize* (1 estimate, #2),; 3) empirical formula from Riley et al., 2020 for MAPE < 0.05 on average (1 estimate, #3); 4) shiny app from Silvey & Liu (2024) for 4 prediction models, LR, XGBoost, Random Forest and Neural Network (4 estimates, #5-8); 5) *samplesizedev* for MAPE < 0.02 on average and for the slope > 0.90 on average and with assurance (3 estimates. #9-11); 6) *pmsims* for slope>0.9 on average and with assurance, and for AUC > $AUC^{max}$-0.02 on average and with assurance (4 estimates, #11-14); 7) results of Kalaycıoğlu et al. (2025) for 6 combinations of the data generators and prediction models (6 estimates, #15-#20); 8) combined results of Ploeg et al. (2014) and Austin et al. (2024) for 3 predictive models (3 estimates, #21-23), 23 estimates total. The 3 case studies (0.063 prevalence, 10 predictors, 0.82 AUC; 0.11 prevalence, 44 predictors, 0.86 AUC; 0.25 prevalence, 17 predictors, 0.80 AUC) represented the datasets used by Kalaycıoğlu et al. (2025) (Cases 1 and 2), and Silvey & Liu (2024). **Colour code**: blue represents estimates for the use of logistic regression as prediction model; red - ML methods (XGBoost, SVM, Random Forest, Neural Network). **Shape code:** circles represent mean criteria, triangles – with assurance, stars – scenarios for the data generator and prediction model mismatch, i.e. misspecified prediction model, e.g. XGBoost for outcome generation and LR for prediction. Black and white triangles show *pmsims* sample size estimate for the AUC reaching AUC*-2 and for the validated calibration slope reaching 0.90 with assurance; grey triangle – calibration slope with assurance by *samplsizedev*. Further details are in the Supplementary Materials, https://github.com/dianashams/pmsims_paper/tree/main.

# 5. Discussion

## 5.1 Taxonomy of sample size methods for prediction models

The development of sample size methods for prediction modelling has been shaped by diverse methodological aims. Methods diverge in the performance characteristics of interest, such as discrimination, calibration, subgroup fairness, or stability, and in assumptions about data richness and model complexity. Most approaches target a minimum level of expected performance; others introduce stricter requirements for achieving the sufficiently performance with assurance. Consequently, closed-form solutions have been developed for classical regression models in simplified scenarios, whereas simulation-based approaches are required for complex, multilevel, or machine-learning settings.

**Figure 4. Taxonomy of sample size calculations**

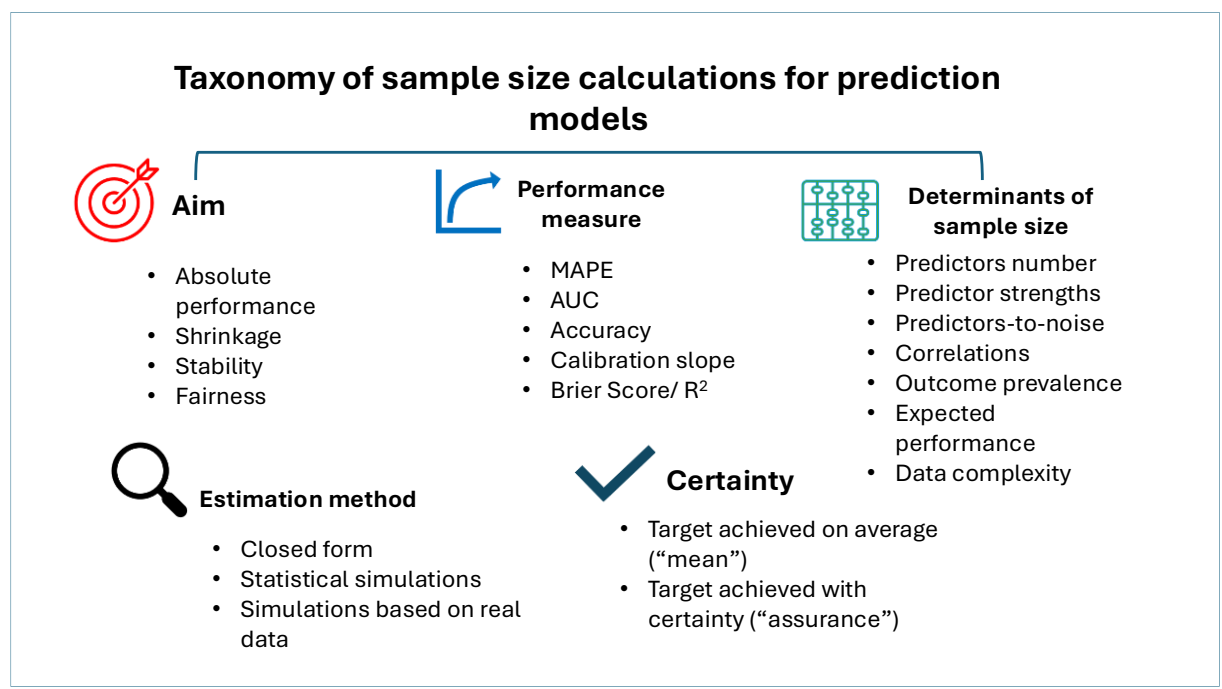

In this paper, we reviewed the methodologies determining development (training) sample size, summarised the theoretical foundations of the problem and illustrated how existing approaches can be applied in practice through case studies. **Figure 4** presents an overview of this methodological landscape. Many existing methods rely on restrictive assumptions about model form, data-generating mechanisms, or performance targets, limiting their applicability in complex or real-world settings. Addressing these limitations, we introduced *pmsims*, a flexible, and customisable simulation-based framework for determining sample size in prediction modelling.

## 5.2 Strengths, limitations and future directions for *pmsims*.

Our approach, *pmsims,* is a simulation-based method that pairs a data generator with a computationally efficient learning curve approximation. The *pmsims* approach solves for the sample size sufficient to achieve a user-defined performance target with an 80% assurance level. The package follows a modular design, allowing it to be used with any model, performance metrics, or data generator, and enabling natural extension to new algorithms and data structures. For ease of use, the package provides functions that address commonly encountered scenarios. The use of Gaussian process surrogate modelling enhances computational efficiency and provides more robust estimates across simulation replications. Further, simulation approach along with the assurance criteria considers model instability in the sample size recommendations. Overall, *pmsims* is compatible with both classical and machine learning models and can be adapted to diverse simulation settings.

Current implementation focuses on standard models (linear, logistic, and Cox), and broader scope of models including ML is available by custom input. Further developments will include expanding data generators to those mimicking clinical data, learning from user-supplied pilot data, and wider range of simulated scenarios. Additional research guiding the users in the optimal number of simulation repetitions to use for complex scenarios is also required. **Table 1** compares *pmsims* with existing frameworks across closed-form, simulation-based, and learning curve paradigms.

## 5.3 Current challenges in the sample size problem for prediction modelling

Despite substantial methodological progress, determining adequate sample sizes for prediction models remains an open and evolving challenge. Even the most recent simulation-based solutions do not readily handle clustered, hierarchical, and longitudinal data that are common in health data and where correlation between observations and temporal dependencies reduce the effective sample size (Riley, Collins, et al., 2025; Tsegaye et al., 2025).

Emerging data modalities, including wearable devices and multi-modal sources combining clinical, imaging, and genomic features, are inherently complex and multi-dimensional. More prediction models are developed on such data given the growing number of digital health tools and data collection platforms (Y. Lu et al., 2024; Raza et al., 2022). This warrants the demand for the sample size estimates in these settings.

Further, investigating sample size requirements for complex data demands data generators that can reflect their structure. Yet most existing tools use simple generators drawing samples from normal or binomial distributions. Specialist data generating packages supporting correlated and hierarchical data through parametric or ML models, such as *simstudy*, *synthpop* (Goldfeld & Wujciak-Jens, 2020; Nowok et al., 2016), and *simsurv* (Brilleman et al., 2021), could be integrated into sample size frameworks. Deep-learning approaches, particularly generative adversarial networks (GANs) (Akiya et al., 2024), also offer a way to generate realistic synthetic data. Key challenges include specifying credible distributional and correlation assumptions for health data and obtaining sufficiently large, realistic datasets to train GANs. Despite these challenges, incorporating advanced data generators could substantially improve the realism and accuracy of sample size estimates in real-world applications. Another substantial methodological gap is incorporating missing data in the sample size requirements. Missing data is almost always present in health data and can substantially impair model performances, yet methodological guidance on sample size adjustments is still to be developed (Getzen et al., 2023).

Finally, there is a growing focus on ensuring that models perform reliably across subgroups and settings, prompting calls to integrate stability and fairness measures into sample size calculations, as well as ensuring stability of individual predictions (Rountree et al., 2024). This is particularly important the field of digital health in the UK, where the NHS specifically targets reduction in health inequality along with an overall treatment benefit (NICE process and methods, 2025).

## 5.4 Conclusion

To conclude, sample size determination remains a central yet underdeveloped aspect of clinical prediction modelling. This paper outlined the conceptual foundations, methodological developments, and current challenges in estimating development sample sizes for reliable model performance. The proposed simulation-based framework *pmsims* offers a flexible and generalisable solution combining learning-curve estimation, Gaussian process optimisation, and assurance-based criteria. Its modular design allows users to tailor data generators, models, and performance metrics to addresses many outstanding challenges. As prediction models increasingly incorporate multimodal and time-series data, future work should establish simulation scenarios that accommodate longitudinal structures, missing data, and fairness considerations, ensuring robust and equitable deployment in clinical and digital health applications.

# Table 1. Comparison of existing methods and software for sample size calculation in prediction modelling.

| Category | Method / Tool | Setting / Model[2] | Performance Focus | Approach | Software availability | Dev/Val[1] | Strengths | Limitations |
|---|---|---|---|---|---|---|---|---|
| *Closed form* | pmsampsize (Ensor et al., 2022) | Linear, Logistic & Cox PH regression | Calibration slope / shrinkage, MAPE | Asymptotic theory informed formulae | CRAN R package, Stata module | Dev | Fast, interpretable, regression-specific | Limited to regression; assumes model correctly specified |
| | Jinks et al. (2015) | Time-to-event (survival) | Royston's D measure | Theory and simulation-informed formulae | Stata command (dsampsi) | Dev | Tailored to survival prognostic models, fast | Narrow scope, experimental assumptions |
| | sampsizeval (Pavlou et al., 2021) | Binary and time-to-event risk models | C-statistic, calibration slope, MAPE | Asymptotic theory informed formulae | CRAN R package, open-source GitHub | Val | Fast, covers sample size for external validation | Limited to binary outcomes |
| *Simulation-based* | **pmsims** | **Linear, logistic and Cox PH, user-defined ML possible** | **C-statistic, calibration slope, user-defined metrics** | **Simulations with Gaussian Process learning curve** | **R package, open-source GitHub** | **Dev** | **Assurance criteria, simple data generators and ML available, ability to add user-defined metrics, data-generators and models** | **Requires coding additional functions to benefit from user-defined data generators and models.** |
| | Silvey & Liu (2024) | Classification algorithms (ML) | C-statistic | Simulations in 16 large health datasets | Shiny app | Dev | Accommodates ML, considers model complexity and data non-linearity | Limited to binary outcomes, heterogeneous datasets |
| | Austin, Lee & Wang (2024) | Logistic regression, RF, bagging, XGBoost | Calibration (slope, intercept, ICI) | Simulations in cardiovascular datasets | No (study recommendations) | Dev | Shows penalized LR less data-hungry than ML | Diverse models, but only 2 datasets |
| | samplesizedev (Pavlou et al., 2025) | Binary and time-to-event risk models | C-statistic, calibration slope, MAPE | Simulation | R package, open-source GitHub | Dev | Simple to use, clear outcome, assurance criteria for calibration slope | Limited to binary outcomes, generic data generator |
| | Kalaycıoğlu et al. (2025) | Logistic regression and tree-based ML models | MAPE | Simulations in cardiovascular datasets with mismatching outcome generators and prediction models | No (study recommendations) | Dev | (First) Exploration of the model misspecification on sample size, ML models | No tool or formula to generalise to other datasets |
| | Goldenholz et al. (2023) | ML validation (Cox PH, ordinal, deep learning) | Precision & accuracy of performance estimates | Simulation/resampling-based | Open-source GitHub | Val | Combines several metrics, model-agnostic and focused on ML | Not a package, requires data preparation |
| *Learning curve* | Figueroa et al. (2012) | Classification (general) | Accuracy / error | Parametric learning curves (inverse power law) | No (conceptual framework) | Dev | Intuitive, performance–n link | Sensitive to curve assumptions |
| | Dayimu et al. (2024) | Prediction models (general) | Discrimination, MAPE | Learning curves - based, utilizes Gaussian Processes | Open-source GitHub | Dev | Clear trajectory-based framework | Relies on availability of pilot and similar external data |

[1] Dev/Val indicate whether the methodologies focus on sample size for developmental (Dev) or validation (Val) datasets.

[2] Abbreviations used in the table: ML – machine learning, MAPE – mean absolute prediction error, Cox PH – Cox Proportionate Hazards model, RF – random forest, SVM – support vector machines, ICI – integrated calibration index.

The table reviews some existing general methodologies and software applications for sample size calculation for prediction modelling, focusing on methods assessing the size of the developmental, or development data. For that reason, only 2 validation sample size studies with available software are mentioned.

# 6. Declarations and acknowledgements

**Funding.** The *pmsims* project, including this study is funded by the NIHR, Research for Patient Benefit programme (NIHR206858). The views expressed are those of the author(s) and not necessarily those of the NIHR or the Department of Health and Social Care. This study is part-funded by the National Institute for Health and Care Research (NIHR) Maudsley Biomedical Research Centre (BRC). DSt, DSh, and GF received financial support by the National Institute for Health Research (NIHR) Biomedical Research Centre at South London and Maudsley NHS Foundation Trust (NIHR203318) and King's College London.

**Ethics approval and consent to participate.** Not applicable.

**Availability of data and materials.** The *pmsims* code is available at https://github.com/pmsims-package/pmsims/.

**Competing interests.** The authors declare that they have no competing interests.

**Authors' contributions.** CRediT authorship contribution statement: DSh: Writing – Original draft, Formal analysis. EC, GF: Funding acquisition. SM, DSh, FZ, DSt, EC, GF: Writing – Review & Editing, Methodology, Conceptualization.

**Acknowledgements.** We are grateful to the Advisory Group of the *pmsims* project for their sustained guidance and support throughout the development of this work. We acknowledge the contributions of Dr Nick Cummins (King's College London, UK) and Dr Joie Ensor (University of Birmingham, UK), whose academic insights strengthened both the conceptual and methodological foundations of the project. We also extend our thanks to public representatives Katherine Barrett, Emma Shellard, and Aurora Todisco, whose lived experience and thoughtful engagement helped ensure that the development of the package remained grounded in real-world needs and user priorities. Their combined expertise significantly enriched the direction and clarity of this work.

# References:

Akiya, I., Ishihara, T., & Yamamoto, K. (2024). Comparison of Synthetic Data Generation Techniques for Control Group Survival Data in Oncology Clinical Trials: Simulation Study. *JMIR Medical Informatics*, *12*, e55118–e55118. https://doi.org/10.2196/55118

Archer, L., Snell, K. I. E., Ensor, J., Hudda, M. T., Collins, G. S., & Riley, R. D. (2021). Minimum sample size for external validation of a clinical prediction model with a continuous outcome. *Statistics in Medicine*, *40*(1), 133–146. https://doi.org/10.1002/SIM.8766

Austin, P. C., Harrell, F. E., & Steyerberg, E. W. (2021). Predictive performance of machine and statistical learning methods: Impact of data-generating processes on external validity in the "large N, small p" setting. *Statistical Methods in Medical Research*, *30*(6), 1465–1483. https://doi.org/10.1177/09622802211002867

Austin, P. C., Lee, D. S., & Wang, B. (2024). The relative data hungriness of unpenalized and penalized logistic regression and ensemble-based machine learning methods: the case of calibration. *Diagnostic and Prognostic Research*, *8*(1), 15. https://doi.org/10.1186/s41512-024-00179-z

Austin, P. C., & Steyerberg, E. W. (2017). Events per variable (EPV) and the relative performance of different strategies for estimating the out-of-sample validity of logistic regression models. *Statistical Methods in Medical Research*, *26*(2), 796–808. https://doi.org/10.1177/0962280214558972

Brilleman, S. L., Wolfe, R., Moreno-Betancur, M., & Crowther, M. J. (2021). Simulating Survival Data Using the **simsurv** R Package. *Journal of Statistical Software*, *97*(3). https://doi.org/10.18637/jss.v097.i03

Carr, E., Forbes, G., Olaniran, R., Shamsutdinova, D., Stahl, D., Markham, S., & Zimmer, F. (2025). *pmsims: Simulation-based Sample Size Tools for Prediction Models*. https://pmsims-package.github.io/pmsims/

Collins, G. S., Moons, K. G. M., Dhiman, P., Riley, R. D., Beam, A. L., Van Calster, B., Ghassemi, M., Liu, X., Reitsma, J. B., van Smeden, M., Boulesteix, A.-L., Camaradou, J. C., Celi, L. A., Denaxas, S., Denniston, A. K., Glocker, B., Golub, R. M., Harvey, H., Heinze, G., … Logullo, P. (2024). TRIPOD+AI statement: updated guidance for reporting clinical prediction models that use regression or machine learning methods. *BMJ*, e078378. https://doi.org/10.1136/bmj-2023-078378

Damen, J. A. A. G., Hooft, L., Schuit, E., Debray, T. P. A., Collins, G. S., Tzoulaki, I., Lassale, C. M., Siontis, G. C. M., Chiocchia, V., Roberts, C., Schlüssel, M. M., Gerry, S., Black, J. A., Heus, P., van der Schouw, Y. T., Peelen, L. M., & Moons, K. G. M. (2016). Prediction models for cardiovascular disease risk in the general population: systematic review. *BMJ*, i2416. https://doi.org/10.1136/bmj.i2416

Dayimu, A., Simidjievski, N., Demiris, N., & Abraham, J. (2024). Sample size determination for prediction models via learning-type curves. *Statistics in Medicine*, *43*(16), 3062–3072. https://doi.org/https://dx.doi.org/10.1002/sim.10121

Dhiman, P., Ma, J., Qi, C., Bullock, G., Sergeant, J. C., Riley, R. D., & Collins, G. S. (2023). Sample size requirements are not being considered in studies developing prediction models for binary outcomes: a systematic review. *BMC Medical Research Methodology*, *23*(1). https://doi.org/10.1186/S12874-023-02008-1

Ensor, J., Martin, E. C., & Riley, R. D. (2022). *pmsampsize: Calculates the Minimum Sample Size Required for Developing a Multivariable Prediction Model*. https://CRAN.R-project.org/package=pmsampsize

Figueroa, R. L., Zeng-Treitler, Q., Kandula, S., & Ngo, L. H. (2012). Predicting sample size required for classification performance. *BMC Medical Informatics and Decision Making*, *12*(1). https://doi.org/10.1186/1472-6947-12-8

Getzen, E., Ungar, L., Mowery, D., Jiang, X., & Long, Q. (2023). Mining for equitable health: Assessing the impact of missing data in electronic health records. *Journal of Biomedical Informatics*, *139*, 104269. https://doi.org/10.1016/j.jbi.2022.104269

Goldenholz, D. M., Sun, H., Ganglberger, W., & Westover, M. B. (2023). Sample Size Analysis for Machine Learning Clinical Validation Studies. *Biomedicines*, *11*(3). https://doi.org/10.3390/BIOMEDICINES11030685

Goldfeld, K., & Wujciak-Jens, J. (2020). simstudy: Illuminating research methods through data generation. *Journal of Open Source Software*, *5*(54), 2763. https://doi.org/10.21105/joss.02763

Kalaycıoğlu, O., Pavlou, M., Akhanlı, S. E., de Belder, M. A., Ambler, G., & Omar, R. Z. (2025). Evaluating the sample size requirements of tree-based ensemble machine learning techniques for clinical risk prediction. *Statistical Methods in Medical Research*. https://doi.org/10.1177/09622802251338983

Kaplan, J., McCandlish, S., Henighan, T., Brown, T. B., Chess, B., Child, R., Gray, S., Radford, A., Wu, J., & Amodei, D. (2020). Scaling laws for neural language models. *ArXiv Preprint ArXiv:2001.08361*.

Lu, C., Reddy, C. K., Wang, P., Nie, D., & Ning, Y. (2024). Multi-Label Clinical Time-Series Generation via Conditional GAN. *IEEE Transactions on Knowledge and Data Engineering*, *36*(4), 1728–1740. https://doi.org/10.1109/TKDE.2023.3310909

Lu, Y., Perumal, T., & Liu, C. (2024). Improved Model for Stroke Prediction on Wearable Devices with CNN_LSTM_KAN. *Proceedings of the 2024 the 12th International Conference on Information Technology (ICIT)*, 225–230. https://doi.org/10.1145/3718391.3718430

NICE process and methods, U. (2025, July 14). *NICE health technology evaluations: the manual. Analysis of data for patient subgroups.* Https://Www.Nice.Org.Uk/Process/Pmg36/Chapter/Economic-Evaluation-2#analysis-of-Data-for-Patient-Subgroups.

Nowok, B., Raab, G. M., & Dibben, C. (2016). **synthpop** : Bespoke Creation of Synthetic Data in *R*. *Journal of Statistical Software*, *74*(11). https://doi.org/10.18637/jss.v074.i11

Ogundimu, E. O., Altman, D. G., & Collins, G. S. (2016). Adequate sample size for developing prediction models is not simply related to events per variable. *Journal of Clinical Epidemiology*, *76*, 175–182. https://doi.org/10.1016/j.jclinepi.2016.02.031

Pate, A., Riley, R. D., Collins, G. S., M, van S., B, V. C., Ensor, J., & Martin, G. P. (2023). Minimum sample size for developing a multivariable prediction model using multinomial logistic regression. *Statistical Methods in Medical Research*, *32*(3), 555–571. https://doi.org/https://dx.doi.org/10.1177/09622802231151220

Pavlou, M. (2021). sampsizeval: Sample Size for Validation of Risk Models with Binary Outcomes. In *CRAN: Contributed Packages*. https://doi.org/10.32614/CRAN.package.sampsizeval

Pavlou, M., Ambler, G., Qu, C., Seaman, S. R., White, I. R., & Omar, R. Z. (2024). An evaluation of sample size requirements for developing risk prediction models with binary outcomes. *BMC Medical Research Methodology*, *24*(1), 146. https://doi.org/https://dx.doi.org/10.1186/s12874-024-02268-5

Pavlou, M., Omar, R. Z., & Ambler, G. (2025). Sample Size Calculations for the Development of Risk Prediction Models that Account for Performance Variability. *ArXiv Preprint ArXiv:2509.14028*.

Peduzzi, P., Concato, J., Kemper, E., Holford, T. R., & Feinstein, A. R. (1996). A simulation study of the number of events per variable in logistic regression analysis. *Journal of Clinical Epidemiology*, *49*(12), 1373–1379. https://doi.org/10.1016/S0895-4356(96)00236-3

Provost, F., Jensen, D., & Oates, T. (1999). Efficient progressive sampling. *Proceedings of the Fifth ACM SIGKDD International Conference on Knowledge Discovery and Data Mining*, 23–32. https://doi.org/10.1145/312129.312188

Raza, M. M., Venkatesh, K. P., & Kvedar, J. C. (2022). Intelligent risk prediction in public health using wearable device data. *Npj Digital Medicine*, *5*(1), 153. https://doi.org/10.1038/s41746-022-00701-x

Riley, R. D., Collins, G. S., Kirton, L., Snell, K. I., Ensor, J., Whittle, R., Dhiman, P., van Smeden, M., Liu, X., Alderman, J., Nirantharakumar, K., Manson-Whitton, J., Westwood, A. J., Cazier, J.-B., Moons, K. G. M., Martin, G. P., Sperrin, M., Denniston, A. K., Harrell, F. E., & Archer, L. (2025). Uncertainty of risk estimates from clinical prediction models: rationale, challenges, and approaches. *BMJ*, e080749. https://doi.org/10.1136/bmj-2024-080749

Riley, R. D., Ensor, J., Snell, K. I. E., Archer, L., Whittle, R., Dhiman, P., Alderman, J., Liu, X., Kirton, L., Manson-Whitton, J., van Smeden, M., Moons, K. G., Nirantharakumar, K., Cazier, J.-B., Denniston, A. K., Van Calster, B., & Collins, G. S. (2025). Importance of sample size on the quality and utility of AI-based prediction models for healthcare. *The Lancet Digital Health*, *7*(6), 100857. https://doi.org/10.1016/j.landig.2025.01.013

Riley, R. D., Ensor, J., Snell, K. I. E., Harrell, F. E., Martin, G. P., Reitsma, J. B., Moons, K. G. M., Collins, G., & van Smeden, M. (2020). Calculating the sample size required for developing a clinical prediction model. *BMJ*, m441. https://doi.org/10.1136/bmj.m441

Riley, R. D., Snell, K. I. E., Ensor, J., Burke, D. L., Harrell, F. E., Moons, K. G. M., & Collins, G. S. (2019a). Minimum sample size for developing a multivariable prediction model: Part I – Continuous outcomes. *Statistics in Medicine*, *38*(7), 1262–1275. https://doi.org/10.1002/SIM.7993

Riley, R. D., Snell, K. I. E., Ensor, J., Burke, D. L., Harrell, F. E., Moons, K. G. M., & Collins, G. S. (2019b). Minimum sample size for developing a multivariable prediction model: PART II - binary and time-to-event outcomes. *Statistics in Medicine*, *38*(7), 1276–1296. https://doi.org/10.1002/SIM.7992

Riley, R. D., Whittle, R., Sadatsafavi, M., Martin, G. P., Pate, A., Collins, G. S., & Ensor, J. (2025). A general sample size framework for developing or updating a clinical prediction model. *ArXiv Preprint ArXiv:2504.18730*.

Rountree, L., Lin, Y.-T., Liu, C., Salvatore, M., Admon, A., Nallamothu, B. K., Singh, K., Basu, A., & Mukherjee, B. (2024). *Reporting of Fairness Metrics in Clinical Risk Prediction Models: A Call for Change*. https://doi.org/10.1101/2024.03.16.24304390

Silvey, S., & Liu, J. (2024). Sample Size Requirements for Popular Classification Algorithms in Tabular Clinical Data: Empirical Study. *Journal of Medical Internet Research*, *26*, e60231. https://doi.org/https://dx.doi.org/10.2196/60231

Steyerberg, E. W. (2019). *Clinical prediction models: A practical approach to development, validation, and updating*. Springer International Publishing.

Tsegaye, B., Snell, K. I. E., Archer, L., Kirtley, S., Riley, R. D., Sperrin, M., Van Calster, B., Collins, G. S., & Dhiman, P. (2025). Larger sample sizes are needed when developing a clinical prediction model using machine learning in oncology: methodological systematic review. *Journal of Clinical Epidemiology*, *180*, 111675. https://doi.org/10.1016/j.jclinepi.2025.111675

van der Ploeg, T., Austin, P. C., & Steyerberg, E. W. (2014). Modern modelling techniques are data hungry: a simulation study for predicting dichotomous endpoints. *BMC Medical Research Methodology*, *14*, 137. https://doi.org/10.1186/1471-2288-14-137

Van Houwelingen, J. C., & Le Cessie, S. (1990). Predictive value of statistical models. *Statistics in Medicine*, *9*(11), 1303–1325. https://doi.org/10.1002/sim.4780091109

van Smeden, M., Moons, K. G., de Groot, J. A., Collins, G. S., Altman, D. G., Eijkemans, M. J., & Reitsma, J. B. (2019). Sample size for binary logistic prediction models: Beyond events per variable criteria. *Statistical Methods in Medical Research*, *28*(8), 2455–2474. https://doi.org/10.1177/0962280218784726

Viering, T., & Loog, M. (2023). The Shape of Learning Curves: A Review. *IEEE Transactions on Pattern Analysis and Machine Intelligence*, *45*(6), 7799–7819. https://doi.org/10.1109/TPAMI.2022.3220744

Vittinghoff, E., & McCulloch, C. E. (2007). Relaxing the rule of ten events per variable in logistic and Cox regression. *American Journal of Epidemiology*, *165*(6), 710–718. https://doi.org/10.1093/aje/kwk052

Wald, A. (1949). Note on the consistency of the maximum likelihood estimate. *The Annals of Mathematical Statistics*, *20*(4), 595–601.

Zimmer, F., & Debelak, R. (2025). Simulation-based design optimization for statistical power: Utilizing machine learning. *Psychological Methods*, *30*(3), 513–536. https://doi.org/10.1037/met0000611

Zimmer, F., Henninger, M., & Debelak, R. (2023). Sample size planning for complex study designs: A tutorial for the mlpwr package. *Behavior Research Methods*, *56*(5), 5246–5263. https://doi.org/10.3758/s13428-023-02269-0